\documentclass[11pt]{article}

\usepackage[final,nohyperref]{acl}

\usepackage{latexsym}
\usepackage[T2A,T1]{fontenc}
\usepackage[utf8]{inputenc}
\usepackage{times}
\usepackage{inconsolata}
\newcommand{\tj}[1]{{\fontencoding{T2A}\fontfamily{cmr}\selectfont #1}}
\DeclareUnicodeCharacter{04E2}{\={\CYRI}}
\DeclareUnicodeCharacter{04E3}{\={\cyri}}
\DeclareUnicodeCharacter{04EE}{\={\CYRU}}
\DeclareUnicodeCharacter{04EF}{\={\cyru}}
\usepackage{microtype}
\usepackage{graphicx}
\usepackage{booktabs}
\usepackage{enumitem}
\setlist{nosep,leftmargin=*}

\newcommand{\BaseLawRten}{49.3}
\newcommand{\BaseLawRone}{23.2}
\newcommand{\BaseLawLo}{46.2}
\newcommand{\BaseLawHi}{52.3}

\newcommand{\BaseTaxRten}{45.9}
\newcommand{\BaseTaxRone}{17.4}
\newcommand{\BaseTaxLo}{36.7}
\newcommand{\BaseTaxHi}{55.0}

\newcommand{\BaseWikiRten}{89.6}
\newcommand{\BaseWikiRone}{75.6}
\newcommand{\BaseWikiLo}{86.8}
\newcommand{\BaseWikiHi}{92.2}

\newcommand{\BaseBelRten}{69.8}
\newcommand{\BaseBelRone}{51.7}
\newcommand{\BaseBelLo}{66.7}
\newcommand{\BaseBelHi}{72.8}

\newcommand{\BaseRuRten}{98.6}
\newcommand{\BaseRuRone}{91.2}
\newcommand{\BaseRuLo}{97.7}
\newcommand{\BaseRuHi}{99.3}
\newcommand{\BaseRuHalf}{0.8}
\newcommand{\BaseEnRten}{99.2}
\newcommand{\BaseEnRone}{93.2}
\newcommand{\BaseEnLo}{98.7}
\newcommand{\BaseEnHi}{99.8}
\newcommand{\BaseEnHalf}{0.6}

\newcommand{\MidLawRten}{87.4}

\newcommand{\MidBelRten}{87.7}

\newcommand{\FtLawRten}{92.4}
\newcommand{\FtLawRone}{68.3}
\newcommand{\FtLawLo}{90.6}
\newcommand{\FtLawHi}{94.0}

\newcommand{\FtTaxRten}{91.7}
\newcommand{\FtTaxRone}{54.1}
\newcommand{\FtTaxLo}{86.2}
\newcommand{\FtTaxHi}{97.2}

\newcommand{\FtWikiRten}{97.2}
\newcommand{\FtWikiRone}{93.8}
\newcommand{\FtWikiLo}{95.8}
\newcommand{\FtWikiHi}{98.6}

\newcommand{\FtBelRten}{92.6}
\newcommand{\FtBelRone}{76.6}
\newcommand{\FtBelLo}{90.9}
\newcommand{\FtBelHi}{94.2}

\newcommand{\FtRuRten}{97.9}
\newcommand{\FtRuRone}{88.2}
\newcommand{\FtRuLo}{96.9}
\newcommand{\FtRuHi}{98.8}

\newcommand{\FtEnRten}{98.1}
\newcommand{\FtEnRone}{90.4}
\newcommand{\FtEnLo}{97.1}
\newcommand{\FtEnHi}{98.9}

\newcommand{\EfiveLawRten}{80.2}

\newcommand{\EfiveTaxRten}{82.6}

\newcommand{\EfiveWikiRten}{96.4}

\newcommand{\EfiveBelRten}{92.2}

\newcommand{\EfiveRuRten}{97.9}

\newcommand{\EfiveEnRten}{98.9}

\newcommand{\EfiveFtLawRten}{93.6}

\newcommand{\EfiveFtTaxRten}{94.5}

\newcommand{\EfiveFtWikiRten}{97.4}

\newcommand{\EfiveFtBelRten}{95.2}
\newcommand{\EfiveFtBelRone}{83.6}

\newcommand{\EfiveFtRuRten}{98.6}

\newcommand{\EfiveFtEnRten}{98.9}

\newcommand{\BgeLawRten}{77.6}

\newcommand{\BgeTaxRten}{78.9}

\newcommand{\BgeWikiRten}{95.0}

\newcommand{\BgeBelRten}{83.8}

\newcommand{\BgeRuRten}{98.9}

\newcommand{\BgeEnRten}{99.0}

\newcommand{\BgeFtLawRten}{91.6}

\newcommand{\BgeFtTaxRten}{92.7}

\newcommand{\BgeFtWikiRten}{97.2}

\newcommand{\BgeFtBelRten}{92.2}
\newcommand{\BgeFtBelRone}{79.3}

\newcommand{\BgeFtRuRten}{98.1}

\newcommand{\BgeFtEnRten}{98.1}

\newcommand{\FtDeltaLawRten}{+43.1}

\newcommand{\FtDeltaLawRone}{+45.1}

\newcommand{\FtDeltaLawOverlap}{separate}

\newcommand{\VsEfiveFtLawRtenAbs}{1.2}

\newcommand{\VsEfiveFtLawRoneAbs}{0.5}
\newcommand{\VsEfiveFtLawOverlap}{overlap}
\newcommand{\VsBgeFtLawRten}{+0.8}

\newcommand{\VsBgeFtLawRoneAbs}{3.0}

\newcommand{\FtDeltaTaxRten}{+45.9}

\newcommand{\FtDeltaTaxRone}{+36.7}

\newcommand{\VsEfiveFtTaxRtenAbs}{2.8}

\newcommand{\VsEfiveFtTaxRoneAbs}{4.6}
\newcommand{\VsEfiveFtTaxOverlap}{overlap}
\newcommand{\VsBgeFtTaxRten}{-0.9}

\newcommand{\VsBgeFtTaxRoneAbs}{8.3}

\newcommand{\FtDeltaWikiRten}{+7.6}

\newcommand{\FtDeltaWikiRone}{+18.2}

\newcommand{\VsEfiveFtWikiRtenAbs}{0.2}

\newcommand{\VsEfiveFtWikiOverlap}{overlap}

\newcommand{\FtDeltaBelRten}{+22.8}

\newcommand{\FtDeltaBelRone}{+24.9}

\newcommand{\VsEfiveFtBelRtenAbs}{2.7}

\newcommand{\VsEfiveFtBelRoneAbs}{7.0}

\newcommand{\FtDeltaRuRten}{-0.7}
\newcommand{\FtDeltaRuRtenAbs}{0.7}
\newcommand{\FtDeltaRuRone}{-3.0}
\newcommand{\FtDeltaRuRoneAbs}{3.0}

\newcommand{\FtDeltaEnRten}{-1.1}
\newcommand{\FtDeltaEnRtenAbs}{1.1}
\newcommand{\FtDeltaEnRone}{-2.8}
\newcommand{\FtDeltaEnRoneAbs}{2.8}

\newcommand{\FtGainFourMean}{+29.9}
\newcommand{\EfiveGainFourMean}{+7.4}
\newcommand{\BgeGainFourMean}{+9.6}
\newcommand{\PctGainAtMidLaw}{88}
\newcommand{\PctGainAtMidBel}{79}
\newcommand{\UnkCount}{0}
\newcommand{\GlyphCount}{12}
\newcommand{\FertTg}{3.01}
\newcommand{\FertXlmrTg}{3.13}

\newcommand{\FertRu}{1.87}

\newcommand{\FertEn}{1.23}

\newcommand{\NLawQ}{1,000}
\newcommand{\NLawCorpus}{72,920}
\newcommand{\NTaxQ}{109}
\newcommand{\NTaxCorpus}{23,282}
\newcommand{\NWikiQ}{500}
\newcommand{\NWikiCorpus}{15,351}
\newcommand{\NBelQ}{900}
\newcommand{\NBelCorpus}{488}
\newcommand{\NRuQ}{900}

\newcommand{\NEnQ}{900}

\newcommand{\NLawHeld}{1,978}
\newcommand{\NLawAll}{19,895}
\newcommand{\NTaxHeld}{22}
\newcommand{\NTaxAll}{233}
\newcommand{\NWikiHeld}{750}

\newcommand{\SamePairs}{identical}
\newcommand{\NPairsTotal}{235,140}
\newcommand{\NPairsLaw}{52,758}
\newcommand{\NPairsLawNeg}{40,259}
\newcommand{\NPairsTax}{1,033}
\newcommand{\NPairsTaxNeg}{875}
\newcommand{\NPairsWiki}{21,123}
\newcommand{\NPairsWikiNeg}{20,374}
\newcommand{\NPairsWeb}{59,985}
\newcommand{\NPairsWebNeg}{39,140}
\newcommand{\NPairsSvc}{15,744}
\newcommand{\NPairsSvcNeg}{6,313}
\newcommand{\NPairsEn}{49,987}
\newcommand{\NPairsRuMiracl}{4,664}
\newcommand{\NPairsRuHnp}{29,846}

\newcommand{\PctReplay}{36}
\newcommand{\NPairsWithNeg}{106,961}
\newcommand{\PctWithNeg}{71}
\newcommand{\TrainSteps}{466}
\newcommand{\TrainLr}{$2\times10^{-5}$}
\newcommand{\TrainMinutes}{166}
\newcommand{\LossFirst}{1.54}
\newcommand{\LossLastMean}{0.35}
\newcommand{\BfNanStep}{3}
\newcommand{\MiniBatch}{16}
\newcommand{\PeakGiB}{18.7}
\newcommand{\NLayers}{24}
\newcommand{\Hidden}{768}
\newcommand{\Vocab}{262,144}
\newcommand{\MaxPos}{2,048}
\newcommand{\Pooling}{mean}
\newcommand{\EmbDim}{768}
\newcommand{\DenseHidden}{3,072}
\newcommand{\QueryPrompt}{task: search result \textbar{} query: }
\newcommand{\DocPrompt}{title: none \textbar{} text: }
\newcommand{\VocabOverXlmr}{1.05}
\newcommand{\BaseFpLawRten}{48.5}

\newcommand{\FpMaxDiff}{0.8}

\newcommand{\ModelName}{Saidzoda-Tajik-EmbeddingGemma-300M}

\title{\ModelName:\\Knowing Every Letter Is Not Reading Tajik}

\author{Tohir Saidzoda \and Nekruz Maxudov \and Bedil Karimov \\
  \textbf{Abubakr Abdullozoda} \and \textbf{Amin Jalilov} \and \textbf{Jonibek Kishvarov} \\
  \textbf{Rahmonali Sharipov} \and \textbf{Mariyamkhon Shodyeva} \\
  Saidzoda AI Lab, Saidzoda Engineering Company, Dushanbe, Tajikistan \\
  \texttt{tohir@saidzoda.com}}

\usepackage{fancyhdr}
\fancypagestyle{plain}{\fancyhf{}\fancyhead[L]{\fontsize{20}{24}\selectfont\textbf{Saidzoda AI Lab}}\fancyfoot[C]{\thepage}}
\begin{document}
\maketitle
\thispagestyle{plain}

\begin{abstract}
EmbeddingGemma-300M is a small open embedder whose tokenizer represents every Tajik letter, and as released it is weak at Tajik retrieval: on our Tajik law test the right passage is in its top ten results for \BaseLawRten\% of questions, on tax documents for \BaseTaxRten\%, and on the human-written Belebele Tajik set for \BaseBelRten\%, against \BaseRuRten\% on the Russian version of the same set.
We present \ModelName, the same model fully fine-tuned for one epoch on \NPairsTotal{} question and passage pairs written by our own Tajik language model for laws, tax documents, Wikipedia and web text, with the same mined hard negatives, English and Russian replay, loss, global batch, learning rate and epoch as our earlier Saidzoda-Tajik-E5-Large (arXiv:2611.00592) and BGE-M3 models.
After training, law reaches \FtLawRten\% and tax \FtTaxRten\%; on Belebele, the one test written by people, recall@10 rises from \BaseBelRten\% to \FtBelRten\% and the first result is right for \FtBelRone\% of questions instead of \BaseBelRone\%.
A checkpoint at step 100 of \TrainSteps{} already holds \PctGainAtMidLaw\% of the law gain at recall@10.
At recall@10 the result is no more than \VsEfiveFtTaxRtenAbs{} points below the two 560M models trained on the same pairs on law, tax and Wikipedia, within the intervals, with \EmbDim-dimensional vectors and about half their parameters; at recall@1, for which we have no intervals, it stays \VsEfiveFtBelRoneAbs{} points behind the E5 model on Belebele.
The cost is a drop of \FtDeltaRuRoneAbs{} and \FtDeltaEnRoneAbs{} points in Russian and English recall@1, larger than for the two earlier models.
We also report that, according to our run notes, a bfloat16 test run produced a non-finite loss at step \BfNanStep{} which the training framework logged as zero, so training was done in fp32.
Our generated test questions come from the same model as the training questions, so we treat those scores as an upper limit until questions written by Tajik speakers exist.

\end{abstract}

\section{Introduction}

Tajik is written in a Cyrillic alphabet with six letters that Russian does not have (\tj{ғ, ӣ, қ, ӯ, ҳ, ҷ}).
In two earlier reports we fine-tuned two 560M multilingual embedders for Tajik retrieval on the same question and passage pairs: multilingual-e5-large-instruct \citep{wang2024me5,saidzoda2026e5} and BGE-M3 \citep{chen2024bgem3,saidzoda2026bge}.
Both share the XLM-RoBERTa tokenizer \citep{conneau2020xlmr}, which maps four Tajik letter forms (\tj{ҷ} in both cases and the capitals \tj{Ӣ} and \tj{Ӯ}) to the unknown token, and the BGE-M3 report found that the recipe still worked.

This report is about the opposite case.
EmbeddingGemma-300M \citep{vera2025embeddinggemma} is a small embedder built on a Gemma 3 backbone with a vocabulary of \Vocab{} tokens that represents all \GlyphCount{} Tajik letter forms.
Nothing in its tokenizer stands between it and Tajik, and yet, as released, it is weak on Tajik domain retrieval: recall@10 of \BaseLawRten{} on our law test and \BaseTaxRten{} on tax, and \BaseBelRten{} on Belebele Tajik against \BaseRuRten{} on Belebele Russian.
It is also the smallest of the three models, with \EmbDim-dimensional vectors instead of 1,024 and about half the parameters, which is our motivation for training it; we did not test it on a device.

We ask three questions.
Does the recipe that worked for the two XLM-RoBERTa models, applied to the same pairs and negatives, recover this model?
How close does a 300M model get to the 560M models trained the same way?
And what does it cost in the replay languages?
Our contributions are:
\begin{enumerate}
\item \textbf{The model and its recipe.} EmbeddingGemma-300M fully fine-tuned for Tajik in one epoch, with the model's own prompts, fp32 and eager attention, and a mid-run checkpoint that shows how fast the gain arrives (Sections~\ref{sec:recipe} and~\ref{sec:results}).
\item \textbf{The other half of the tokenizer question.} Full letter coverage, zero-shot collapse, recovery by fine-tuning: coverage is a precondition for learning a language, not a predictor of how a released model does on it (Section~\ref{sec:coverage}).
\item \textbf{A three-way comparison on the same data.} Three embedders, the same pairs and negatives, the same loss, global batch, learning rate and epoch, one frozen test, with the remaining differences and the costs stated (Section~\ref{sec:results}).
\item \textbf{A logging trap, seen in the run notes.} In a bfloat16 test run the loss became non-finite at step \BfNanStep{} and the framework logged it as zero; the earlier reports could only suspect this mechanism (Section~\ref{sec:negative}).
\end{enumerate}

\section{Related Work}

\paragraph{Small multilingual embedders.}
EmbeddingGemma \citep{vera2025embeddinggemma} is a 300M encoder derived from Gemma 3, trained with task prompts for retrieval, classification and similarity, with a \EmbDim-dimensional output that can be truncated.
Other embedders built on decoder language models include Qwen3-Embedding \citep{zhang2025qwen3emb} and the Giga-Embeddings family \citep{kolodin2025gigaembeddings}; the XLM-RoBERTa family includes multilingual E5 \citep{wang2024me5}, BGE-M3 \citep{chen2024bgem3} and mGTE \citep{zhang2024mgte}.
Our E5 report \citep{saidzoda2026e5} compared nine of these models on the Tajik test used here, used as released, and found that the small models were the weakest on Tajik law; this report takes the smallest of them and trains it.
MTEB and MMTEB \citep{muennighoff2023mteb,enevoldsen2025mmteb} include Belebele used as retrieval for Tajik, which we also use, but no Tajik domain task.

\paragraph{Training.}
We fine-tune with the in-batch multiple-negatives loss \citep{henderson2017smartreply} with gradient caching \citep{gao2021gradcache}, following Sentence-BERT \citep{reimers2019sbert}, and mine hard negatives with the positive-aware margin of NV-Retriever \citep{moreira2024nvretriever}.
Training retrievers on queries written by a language model goes back to InPars, Promptagator and GPL \citep{bonifacio2022inpars,dai2022promptagator,wang2022gpl}; whether test sets built the same way favour systems close to the generator is debated \citep{rahmani2024synthetic,rahmani2025bias}, so we read our generated test sets as upper bounds and rely on Belebele \citep{bandarkar2024belebele} for an independent check.

\paragraph{Tajik.}
Tajik has recently gained web corpora \citep{arabov2026tajikweb} and NLP resources \citep{arabov2026tajiknlp}; the closest related retrieval work is on Persian \citep{zinvandi2025famteb,sarmadi2025hakim}, written in a different script. Our own earlier reports adapted language models to Tajik: a 2.6B hybrid model for tool calling \citep{saidzoda2026lfm} and the Qwen3 family from 0.6B to 8B \citep{saidzoda2026qwen3}.
Apart from our own reports, we found no published Tajik embedder.

\section{Every Letter, No Retrieval}
\label{sec:coverage}

We passed each of the \GlyphCount{} Tajik-specific letter forms (the six letters in both cases) through the tokenizer on its own and checked whether the unknown-token id appeared among the token ids.
It never did: \UnkCount{} of \GlyphCount{} forms are unknown, against four for the XLM-RoBERTa tokenizer of our two earlier models.
The vocabulary is also slightly larger (\Vocab{} tokens, \VocabOverXlmr{} times XLM-RoBERTa's), and it splits a Tajik word into \FertTg{} tokens on average on our law chunks, about the same as XLM-RoBERTa's \FertXlmrTg, against \FertRu{} for Russian and \FertEn{} for English on Belebele passages.

Coverage did not translate into retrieval.
Used as released, with its own query and document prompts, the model finds the right law chunk in its top ten for \BaseLawRten\% of questions and the right tax chunk for \BaseTaxRten\%, with recall@1 of \BaseLawRone{} and \BaseTaxRone{}; on Belebele Tajik it reaches \BaseBelRten{} at recall@10 and \BaseBelRone{} at recall@1.
On the Russian and English versions of Belebele the same model scores \BaseRuRten{} and \BaseEnRten{}.
So the model can read Tajik characters, and tokenises Tajik no worse than the XLM-RoBERTa models, but as released it has not learned what Tajik questions and passages have in common.
The two XLM-RoBERTa models, with four Tajik letter forms missing, start from \BgeLawRten{} and \EfiveLawRten{} on law.
Read together with the BGE-M3 report \citep{saidzoda2026bge}, the two observations are these: full letter coverage did not predict zero-shot Tajik retrieval, and the loss of four letter forms did not prevent recovery by fine-tuning.
They come from three models that differ in size and pretraining as well as in tokenizer, so we state them as observations, not as a rule.

\section{Data and Benchmark}
\label{sec:data}

The training pairs, the held-out split and the test sets are exactly those of our E5 report \citep{saidzoda2026e5}; the pair counts of the three training runs are \SamePairs{} (Table~\ref{tab:data}).
We summarise them so that this report can be read alone.

\paragraph{Sources and split.}
All Tajik passages come from four collections: Tajik laws split into \NLawCorpus{} chunks, tax-code and tax-guidance documents split into \NTaxCorpus{} chunks, Tajik Wikipedia, and cleaned Tajik web and literature text.
For each passage our Tajik fine-tune of Qwen3.6-27B wrote questions that the passage answers, and a second, independent call of the model checked each question against its passage; only checked questions were kept.
Before any training pair was built, test material was frozen at the level of whole units, chosen by the SHA-1 hash of the unit's title: \NLawHeld{} of \NLawAll{} laws, \NTaxHeld{} of \NTaxAll{} tax documents and \NWikiHeld{} Wikipedia articles.
No passage from a held-out unit is used in training or as a negative, and any training query whose normalised text equals a test query is dropped.

\paragraph{Test sets.}
\begin{itemize}
\item \textbf{Law}: \NLawQ{} questions on held-out laws, searched over all \NLawCorpus{} law chunks.
\item \textbf{Tax}: \NTaxQ{} questions on held-out documents, over all \NTaxCorpus{} tax chunks. The set is small and its intervals are wide.
\item \textbf{Wikipedia}: \NWikiQ{} questions on held-out articles, over \NWikiCorpus{} passages.
\item \textbf{Belebele} \citep{bandarkar2024belebele}: the \NBelQ{} human-written Tajik questions matched to their passages among \NBelCorpus{} distinct passages, as in the Belebele retrieval task of MMTEB \citep{enevoldsen2025mmteb}, reported as recall rather than nDCG. This is the only test set not produced by our generator. The Russian and English versions check that those languages are kept.
\end{itemize}
The main metric is recall@10; we also report recall@1.
Each recall comes with a 95\% bootstrap interval over questions (1,000 resamples), with the passages held fixed.
All models are scored with the same script (sentence-transformers 5.7), loaded in bfloat16 (an fp32 re-score of the two EmbeddingGemma models changes no recall@10 by more than \FpMaxDiff{} points; see Limitations), each with its own prompts: EmbeddingGemma's query prompt ``\QueryPrompt'' and document prompt ``\DocPrompt'', the E5 query instruction for the E5 models, and none for BGE-M3.
The held-out lists and the test questions are available on request.

\paragraph{Training pairs.}
Table~\ref{tab:data} lists the pairs by domain.
A further \NPairsSvc{} pairs map short customer-service requests, written by a language model in a simulation of our own customer-service line, to topic descriptions; this set is not released.
\PctReplay\% of the pairs are replay: English Natural Questions \citep{kwiatkowski2019nq}, Russian MIRACL \citep{zhang2023miracl} and the Russian hard-negative set ru-HNP.

\begin{table}[t]\centering\footnotesize\setlength{\tabcolsep}{4pt}
\begin{tabular}{lrr}
\toprule
Domain & Pairs & With negative \\
\midrule
Tajik law & \NPairsLaw & \NPairsLawNeg \\
Tajik tax & \NPairsTax & \NPairsTaxNeg \\
Tajik Wikipedia & \NPairsWiki & \NPairsWikiNeg \\
Tajik web and literature & \NPairsWeb & \NPairsWebNeg \\
Customer service (synthetic) & \NPairsSvc & \NPairsSvcNeg \\
\midrule
English replay (NQ) & \NPairsEn & n/a \\
Russian replay (MIRACL) & \NPairsRuMiracl & n/a \\
Russian replay (HNP set) & \NPairsRuHnp & n/a \\
\midrule
Total & \NPairsTotal & \NPairsWithNeg \\
\bottomrule
\end{tabular}
\caption{Training pairs by domain, as counted by the training run; identical to the E5 and BGE-M3 runs of our earlier reports. Replay pairs use in-batch negatives only.}
\label{tab:data}
\end{table}

\section{Training Recipe}
\label{sec:recipe}

\paragraph{Starting model.}
We start from EmbeddingGemma-300M \citep{vera2025embeddinggemma}: a Gemma 3 text encoder with \NLayers{} layers and hidden size \Hidden, vocabulary \Vocab, context \MaxPos{} tokens, \Pooling{} pooling followed by two linear layers (\Hidden{} to \DenseHidden{} to \EmbDim) and normalisation, giving \EmbDim-dimensional vectors.
Its query prompt ``\QueryPrompt'' and document prompt ``\DocPrompt'' are kept in training and at use time, on the anchor and on the positive and negative passages respectively, exactly as at inference.

\paragraph{Hard negatives.}
The negatives are those mined once for the E5 run and shared by all three runs: for every in-domain pair, the passage pool of its domain was searched with the released BGE-M3 and one negative was taken at random (seed 7) from ranks 10 to 50, from a different law, document or article than the positive, not equal to the positive, and with a score below 0.95 times the score of the positive \citep{moreira2024nvretriever}.
\NPairsWithNeg{} pairs (\PctWithNeg\% of the in-domain pairs) received a negative; the rest, and all replay pairs, train with in-batch negatives only.

\paragraph{Loss and settings.}
We train with the cached multiple-negatives ranking loss of sentence-transformers (in-batch negatives plus the mined negative, scale 20), with a global batch of 512 on one H100 GPU shared with another job.
Each batch is drawn from a single domain, domains are sampled in proportion to their size, and a batch never contains the same text twice.
The learning rate is \TrainLr{} with a linear schedule and 5\% warm-up, for one epoch (\TrainSteps{} steps), with sequences up to 512 tokens and seed 42.
Three settings differ from the two XLM-RoBERTa runs, and none affects the global batch: the model's own prompts are used instead of the E5 instruction or no prompt; the gradient-caching mini-batch is \MiniBatch{} instead of 64, for memory; and attention runs in eager mode, chosen together with fp32 after a bfloat16 test run failed (Section~\ref{sec:negative}), so the two were not separated.
As in the earlier runs, training is in fp32 with TF32 matrix multiplication, with the AdamW optimiser and the default weight decay of the trainer; the fp32 run peaked at \PeakGiB{} GiB of GPU memory.
Library versions recorded with the model: sentence-transformers 5.7.0, transformers 4.57.6, PyTorch 2.10.0.
The run took about \TrainMinutes{} minutes on the shared card, three times the E5 run on a free card, so we do not read anything into the time.
A checkpoint was kept every 100 steps; we evaluate the one at step 100 beside the final model.

\begin{figure}[t]\centering
\includegraphics[width=0.9\columnwidth]{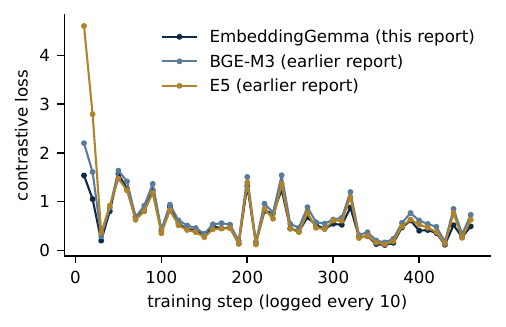}
\caption{Training loss of \ModelName{} and of the E5 and BGE-M3 models of our earlier reports, all on the same pairs with the same settings, logged every 10 steps.}
\label{fig:loss}
\end{figure}

\paragraph{Training loss.}
The logged loss, a single batch every 10 steps and therefore noisy and domain-dependent, falls from \LossFirst{} at step 10 to a mean of \LossLastMean{} over the last five logged steps (Figure~\ref{fig:loss}); the E5 and BGE-M3 runs on the same pairs are drawn beside it.
The three curves are not comparable in absolute value, since each model starts from its own similarity scale and prompt, but all three flatten within the first epoch.

\section{Results}
\label{sec:results}

\begin{table*}[t]\centering\small
\begin{tabular}{lrrr}
\toprule
Test set (queries) & EmbeddingGemma-300M & Ours & Change \\
\midrule
Law (\NLawQ) & \BaseLawRten\ {\scriptsize[\BaseLawLo, \BaseLawHi]} & \FtLawRten\ {\scriptsize[\FtLawLo, \FtLawHi]} & \FtDeltaLawRten \\
Tax (\NTaxQ) & \BaseTaxRten\ {\scriptsize[\BaseTaxLo, \BaseTaxHi]} & \FtTaxRten\ {\scriptsize[\FtTaxLo, \FtTaxHi]} & \FtDeltaTaxRten \\
Wikipedia (\NWikiQ) & \BaseWikiRten\ {\scriptsize[\BaseWikiLo, \BaseWikiHi]} & \FtWikiRten\ {\scriptsize[\FtWikiLo, \FtWikiHi]} & \FtDeltaWikiRten \\
Belebele Tajik (\NBelQ) & \BaseBelRten\ {\scriptsize[\BaseBelLo, \BaseBelHi]} & \FtBelRten\ {\scriptsize[\FtBelLo, \FtBelHi]} & \FtDeltaBelRten \\
Belebele Russian (\NRuQ) & \BaseRuRten\ {\scriptsize[\BaseRuLo, \BaseRuHi]} & \FtRuRten\ {\scriptsize[\FtRuLo, \FtRuHi]} & \FtDeltaRuRten \\
Belebele English (\NEnQ) & \BaseEnRten\ {\scriptsize[\BaseEnLo, \BaseEnHi]} & \FtEnRten\ {\scriptsize[\FtEnLo, \FtEnHi]} & \FtDeltaEnRten \\
\midrule
\multicolumn{4}{l}{\emph{Recall@1}} \\
Law & \BaseLawRone & \FtLawRone & \FtDeltaLawRone \\
Tax & \BaseTaxRone & \FtTaxRone & \FtDeltaTaxRone \\
Wikipedia & \BaseWikiRone & \FtWikiRone & \FtDeltaWikiRone \\
Belebele Tajik & \BaseBelRone & \FtBelRone & \FtDeltaBelRone \\
Belebele Russian & \BaseRuRone & \FtRuRone & \FtDeltaRuRone \\
Belebele English & \BaseEnRone & \FtEnRone & \FtDeltaEnRone \\
\bottomrule
\end{tabular}
\caption{Recall@10 (top block, with 95\% bootstrap intervals) and recall@1 (bottom block) in percent, for EmbeddingGemma-300M as released and after our fine-tuning. Change is in percentage points. Law, tax and Wikipedia queries were written by an LLM on held-out documents; only the Belebele sets are human-written.}
\label{tab:main}
\end{table*}

\paragraph{Gains over the released model.}
Table~\ref{tab:main} compares \ModelName{} with EmbeddingGemma-300M as released.
On law, recall@10 rises from \BaseLawRten{} to \FtLawRten{} (\FtDeltaLawRten{} points) and recall@1 from \BaseLawRone{} to \FtLawRone; on tax, recall@10 rises by \FtDeltaTaxRten{} points.
Wikipedia, where the released model was already usable, rises by \FtDeltaWikiRten{} points at recall@10 and \FtDeltaWikiRone{} at recall@1.
The recall@10 intervals are \FtDeltaLawOverlap{} on every one of these sets; the result files hold no intervals for recall@1, so all recall@1 figures in this report are point estimates.
These three sets were produced by the same generator as the training queries, so their absolute values are optimistic.

On Belebele Tajik, which is human-written and was not used to build any training pair, recall@10 rises by \FtDeltaBelRten{} points, from \BaseBelRten{} to \FtBelRten, with recall@10 intervals that do not overlap, and recall@1 by \FtDeltaBelRone{} points, from \BaseBelRone{} to \FtBelRone.
Russian and English recall@10 fall by \FtDeltaRuRtenAbs{} and \FtDeltaEnRtenAbs{} points, with overlapping intervals (half-widths \BaseRuHalf{} and \BaseEnHalf; no paired test), but recall@1 falls by \FtDeltaRuRoneAbs{} and \FtDeltaEnRoneAbs{} points, from \BaseRuRone{} and \BaseEnRone{} to \FtRuRone{} and \FtEnRone.
As point estimates these are the largest replay costs of the three models we trained on these pairs; the E5 model lost nothing at recall@10 and the BGE-M3 model under one point.
We did not train without replay, so we cannot say how much larger the cost would be without it.

\paragraph{How fast the gain arrives.}
The checkpoint at step 100 of \TrainSteps{} already scores \MidLawRten{} on law and \MidBelRten{} on Belebele Tajik at recall@10, that is \PctGainAtMidLaw\% of the final law gain and \PctGainAtMidBel\% of the final Belebele gain at recall@10 (Table~\ref{tab:compare}).
At that step the learning rate is still near its peak, so this checkpoint is not what a short schedule would produce; the rest of the epoch adds the remaining gain.
We did not evaluate later checkpoints or a second epoch, and we have one seed.

\begin{table*}[t]\centering\small
\begin{tabular}{llrrrrrrr}
\toprule
Model & Size & Used as & Law & Tax & Wiki & Bel.\ tg & Bel.\ ru & Bel.\ en \\
\midrule
EmbeddingGemma-300M & 0.31B & zero-shot & \BaseLawRten & \BaseTaxRten & \BaseWikiRten & \BaseBelRten & \BaseRuRten & \BaseEnRten \\
Ours, step 100 & 0.31B & full FT & \MidLawRten & n/a & n/a & \MidBelRten & n/a & n/a \\
Ours (EmbeddingGemma, fine-tuned) & 0.31B & full FT & \FtLawRten & \FtTaxRten & \FtWikiRten & \FtBelRten & \FtRuRten & \FtEnRten \\
\midrule
BGE-M3 & 0.57B & zero-shot & \BgeLawRten & \BgeTaxRten & \BgeWikiRten & \BgeBelRten & \BgeRuRten & \BgeEnRten \\
Saidzoda-Tajik-BGE-M3 & 0.57B & full FT & \BgeFtLawRten & \BgeFtTaxRten & \BgeFtWikiRten & \BgeFtBelRten & \BgeFtRuRten & \BgeFtEnRten \\
\midrule
multilingual-e5-large-instruct & 0.56B & zero-shot & \EfiveLawRten & \EfiveTaxRten & \EfiveWikiRten & \EfiveBelRten & \EfiveRuRten & \EfiveEnRten \\
Saidzoda-Tajik-E5-Large & 0.56B & full FT & \textbf{\EfiveFtLawRten} & \textbf{\EfiveFtTaxRten} & \textbf{\EfiveFtWikiRten} & \textbf{\EfiveFtBelRten} & \textbf{\EfiveFtRuRten} & \textbf{\EfiveFtEnRten} \\
\bottomrule
\end{tabular}
\caption{Recall@10 (percent) on the same frozen test sets for the three embedders we fine-tuned on the same pairs and negatives with the same loss, global batch, learning rate and epoch, each as released and after fine-tuning; the step-100 row is a checkpoint of our run. Bold marks the highest value among the three fine-tuned models in each column, not a significant difference. The BGE-M3 and E5 rows are the numbers of our earlier reports, re-read from the same result files.}
\label{tab:compare}
\end{table*}

\paragraph{Against the two 560M models trained the same way.}
Table~\ref{tab:compare} shows the three embedders we have fine-tuned on the same pairs and negatives with the same loss, global batch, learning rate and epoch, each as released and after training; the settings that differ are listed in Section~\ref{sec:recipe}.
EmbeddingGemma starts far below the other two on every Tajik set (law \BaseLawRten{} against \BgeLawRten{} and \EfiveLawRten) and gains the most (\FtGainFourMean{} points on the mean of the four Tajik sets, against \BgeGainFourMean{} for BGE-M3 and \EfiveGainFourMean{} for E5).
After training, at recall@10, it is \VsEfiveFtLawRtenAbs{} points below the E5 model on law, \VsEfiveFtTaxRtenAbs{} on tax and \VsEfiveFtWikiRtenAbs{} on Wikipedia, all with overlapping intervals (\VsEfiveFtLawOverlap, \VsEfiveFtTaxOverlap, \VsEfiveFtWikiOverlap), and level with the BGE-M3 model (\VsBgeFtLawRten{} on law, \VsBgeFtTaxRten{} on tax).
At recall@1 the picture is less even: on law it is \VsBgeFtLawRoneAbs{} points above BGE-M3 and \VsEfiveFtLawRoneAbs{} below E5, but on tax (\NTaxQ{} questions) it is \VsEfiveFtTaxRoneAbs{} below E5 and \VsBgeFtTaxRoneAbs{} below BGE-M3.
On Belebele Tajik it ends \VsEfiveFtBelRtenAbs{} points below E5 at recall@10, with overlapping intervals, but \VsEfiveFtBelRoneAbs{} points below at recall@1 (\FtBelRone{} against \EfiveFtBelRone), and at recall@1 it is also below BGE-M3 (\BgeFtBelRone).
So at recall@10 on the in-domain tests the 300M model reaches the two 560M models within the noise; at recall@1, and on the human-written reading questions, a gap remains.
We did not test what causes the remaining gap; the smaller hidden size and the different pretraining are both candidates.

\paragraph{Size.}
The model produces \EmbDim-dimensional vectors against 1,024 for the two XLM-RoBERTa models, so an index of the same passages takes three quarters of the space, and it has about half their parameters.
We do not report latency for this run: it was measured on a card shared with another job and is not comparable with the earlier reports.

\section{What Did Not Work}
\label{sec:negative}

\paragraph{bfloat16 training, and a loss logged as zero.}
Our first short test run of this recipe on EmbeddingGemma used bfloat16.
According to our run notes, at step \BfNanStep{} the loss became non-finite and the training framework, with its default filter for non-finite values, logged it as 0.0; the test run's log itself is not part of the released records, so we report this as noted at the time.
Our E5 report had seen the same zero in two test runs on the XLM-RoBERTa backbone and could only suspect this mechanism.
We then disabled the filter, so that a non-finite loss is written as such and a guard can stop the run, and switched to fp32 and eager attention together, which trained normally; we did not test the two changes separately, nor whether the attention mode of the failed run mattered.
We recommend both settings for any run on this backbone: turn the filter off, and treat a logged loss of exactly zero as a failure.

\paragraph{Replay did not fully protect Russian and English.}
With the same \PctReplay\% of replay pairs that kept the E5 model's Russian and English unchanged, this model lost \FtDeltaRuRoneAbs{} and \FtDeltaEnRoneAbs{} points of recall@1 in those languages (Table~\ref{tab:main}).
We have one seed, no intervals for recall@1 and no run without replay, so we cannot separate the effect of the replay share from that of the model; a larger replay share or a lower learning rate are the obvious things to try.

\section{Release}
\label{sec:release}

\paragraph{What we release.}
The model is on the Hugging Face Hub as \texttt{Tohirju/\allowbreak\ModelName} in sentence-transformers format with its prompts, together with the training log, the step-100 checkpoint's scores and the result files behind the scores in this report.
Access is given on request: the repository is public, and each request is approved by us.
The training pairs, the code and the full test questions are available on request; Table~\ref{tab:examples} shows examples of the test questions.
The customer-service pairs used in training are not released.

\paragraph{Licence.}
The starting model, EmbeddingGemma-300M, is distributed under the Gemma Terms of Use (\texttt{ai.google.dev/gemma/terms}), and our fine-tuned weights are a model derivative under those terms: the repository states that the weights were modified, links the Terms and the Gemma Prohibited Use Policy, and passes their use restrictions on to every user, and the gate requires accepting them.
In addition, our fine-tuned model is offered under the Saidzoda model licence: organisations with revenue need a paid licence, while startups and individuals without revenue may use it free of charge until they earn from it; where the two conflict, the Gemma Terms govern. Access requests go through \texttt{ai.saidzoda.com/access}.

\paragraph{How to use it.}
Load the model with sentence-transformers, encode questions with the ``query'' prompt and passages with the ``document'' prompt, as for the released EmbeddingGemma, and rank passages by cosine similarity.
The model was trained and tested with inputs of up to 512 tokens.

\paragraph{Intended use.}
The model is meant for finding Tajik passages that answer a Tajik question, for example as the first step of a question-answering system over laws or official documents; its size makes it a candidate for small servers and devices, which we did not test.
The retrieved text should be checked before an answer reaches a user.
We report no test on spoken or informal Tajik.

\section{Conclusion}

A tokenizer that knows every Tajik letter left EmbeddingGemma-300M weak on Tajik law and tax as released; one epoch on \NPairsTotal{} pairs brought it, at recall@10, within the noise of two models nearly twice its size on the in-domain tests, and a checkpoint after 100 steps already had \PctGainAtMidLaw\% of the law gain.
With the BGE-M3 report \citep{saidzoda2026bge} this gives two observations that point the same way: a tokenizer's letter coverage did not predict how a released model did on Tajik, in either direction.
The 300M model is the candidate when vectors must be small or the hardware is modest; the costs are \VsEfiveFtBelRoneAbs{} points on human-written questions at recall@1 against the E5 model and a larger loss of Russian and English first-hit rate, which a replay sweep could reduce and a second seed would put an interval on.
The missing evaluation is the same as in our earlier reports: questions written by Tajik speakers about real laws and tax rules.

\section*{Limitations}

\paragraph{Same generator for training and test queries.}
The law, tax and Wikipedia questions in both training and test were written by the same model, on disjoint laws, documents and articles, so a retriever can learn the generator's question style and these scores are upper bounds.
Belebele is independent, but it consists of reading-comprehension questions about short news and travel texts, not the questions people ask about laws.

\paragraph{Statistics.}
The intervals are per-model bootstrap intervals; we did not store per-question ranks and so could not run paired tests.
The tax set has only \NTaxQ{} questions.
The model was trained once, with one seed; the recall@10 differences between it and the two earlier models have overlapping intervals on every in-domain set, and recall@1 has no intervals at all in our result files.

\paragraph{Confounds in the three-way comparison.}
The three runs share pairs, negatives, loss, global batch, learning rate, epoch and precision, but not everything: this run used the model's own prompts, a smaller gradient-caching mini-batch and eager attention, and ran on a shared card.
None of these changes the global batch, but we did not test them one at a time.
The hard negatives were mined with the released BGE-M3, which may give the BGE-M3 fine-tune a small edge in that comparison.
Only one intermediate checkpoint was evaluated, and only on the Tajik sets.

\paragraph{Scoring precision.}
All headline scores are in bfloat16.
Because the same model produced a non-finite loss when trained in bfloat16, we re-scored both the released EmbeddingGemma and our fine-tune in fp32 on all six sets.
No recall@10 value moved by more than \FpMaxDiff{} points, and the released model's law score was \BaseFpLawRten{} in fp32 against \BaseLawRten{} in bfloat16, so its low zero-shot score on Tajik is not a precision effect.

\paragraph{Scope.}
We tested retrieval only, not reranking or end-to-end question answering, and only text up to 512 tokens; the model's output can be truncated to fewer dimensions, which we did not evaluate.
The customer-service training pairs come from a simulation and are not released, so that part of the training data cannot be reproduced by others.
No latency is reported for this run.

\section*{Use of AI Assistants}

We used AI assistants (Anthropic Claude models) to write data-processing, training and analysis scripts, to search for and check references, and to draft parts of the text.
The training questions were generated by our own Tajik language model, as described in Section~\ref{sec:data}.
The authors designed the study, ran the training, checked every number against the recorded files, and take full responsibility for the content.
Every number in this report is generated by a script from logged files, and each reference was checked against its arXiv, ACL Anthology or DOI record.
AI assistants are not authors.

\section*{Acknowledgements}

We thank NVIDIA for access to the NVIDIA Inception programme, through which the training compute for this work was provided.

\bibliography{refs}
\bibliographystyle{acl_natbib}

\appendix
\section{Example Test Questions}
\begin{table*}[t]\centering\small
\begin{tabular}{lp{0.43\textwidth}p{0.38\textwidth}}
\toprule
Set & Question (Tajik) & English gloss \\
\midrule
Law & \tj{Мувофиқи моддаи} 43-\tj{и Кодекси андози Тоҷикистон, объектҳои муоинаи хронометражӣ кадомҳоянд}? & Under Article 43 of the Tax Code of Tajikistan, what are the objects of a timekeeping inspection? \\
Law & \tj{Тибқи Моддаи} 346-\tj{и Кодекси андози Ҷумҳурии Тоҷикистон, манбаи андоз чӣ тавр муайян карда мешавад}? & Under Article 346 of the Tax Code, how is the tax base determined? \\
Tax & \tj{Муқаррароти Кодекси одоби касбии мақомоти андоз ба кадом категорияи кормандон татбиқ мегардад}? & To which categories of staff do the rules of the tax authorities' code of professional ethics apply? \\
Tax & \tj{Корманди андоз бояд чӣ гуна амал кунад, то ба ошкоркунии сирри давлатӣ роҳ надиҳад}? & How must a tax officer act so as not to disclose state secrets? \\
Wikipedia & \tj{Дар замони Имперотурии Юан ҳокимияти имперотурӣ дар ҳар як вилоят аз ҷониби кадом намояндагон ташкил карда мешуд}? & In the Yuan Empire, which representatives exercised imperial authority in each province? \\
Wikipedia & \tj{Чаро дар интихоботи соли} 1924 \tj{Ҳизби Миллӣ ва ҳизби Меҳнатӣ пирӯз шуданд}? & Why did the National Party and the Labour Party win the 1924 election? \\
\bottomrule
\end{tabular}
\caption{Example test questions from the frozen law, tax and Wikipedia sets, exactly as written by the generator, without editing. The English glosses are ours. The full sets are available on request.}
\label{tab:examples}
\end{table*}

\end{document}